\def\year{2020}\relax
\documentclass[letterpaper]{article} 
\usepackage{aaai20}  
\usepackage{times}  
\usepackage{helvet} 
\usepackage{courier}  
\usepackage[hyphens]{url}  
\usepackage{graphicx} 
\usepackage{subfigure}
\usepackage{multirow}
\usepackage{amsmath}
\usepackage{adjustbox}
\usepackage{threeparttable}
\usepackage{graphicx}  
\usepackage{color}

\title{Multi-Label Classification with Label Graph Superimposing}
\author{
Ya Wang$^{\$}\thanks{equal contribution. This work was done when Ya Wang was a full-time research intern at Baidu.}$,
Dongliang He$^{\ddag*}$,
Fu Li$^{\ddag}$, 
Xiang Long$^{\ddag}$,
Zhichao Zhou$^{\ddag}$,
Jinwen Ma$^{\$}\thanks{Corresponding author}$,
Shilei Wen$^{\ddag}$
\\
$^{\$}$School of Mathematical Sciences and LMAM, Peking University, China\\
$^{\ddag}$Department of Computer Vision Technology (VIS), Baidu Inc., Beijing, China\\
\{wangyachn@, jwma@math\}.pku.edu.cn ~~\{hedongliang01, lifu, longxiang, zhouzhichao01, wenshilei\}@baidu.com
}

\begin{document}

\maketitle

\begin{abstract}
Images or videos always contain multiple objects or actions. Multi-label recognition has been witnessed to achieve pretty performance attribute to the rapid development of deep learning technologies. Recently, graph convolution network (GCN) is leveraged to boost the performance of multi-label recognition. However, what is the best way for label correlation modeling and how feature learning can be improved with label system awareness are still unclear. In this paper, we propose a label graph superimposing framework to improve the conventional GCN+CNN framework developed for multi-label recognition in the following two aspects. Firstly, we model the label correlations by  superimposing label graph built from statistical co-occurrence information into the graph constructed from knowledge priors of labels, and then multi-layer graph convolutions are applied on the final superimposed graph for label embedding abstraction.  Secondly, we propose to leverage embedding of the whole label system for better representation learning. In detail, lateral connections between GCN and CNN are added at shallow, middle and deep layers to inject information of label system into backbone CNN for label-awareness in the feature learning process. 
Extensive experiments are carried out on MS-COCO and Charades datasets, showing that our proposed solution can greatly improve the recognition performance and achieves new state-of-the-art recognition performance. 
\end{abstract}

\section{Introduction}

\begin{figure}[!t]
	\centering
	\subfigure[Examples on MS-COCO]  {\includegraphics[width=0.95\columnwidth]{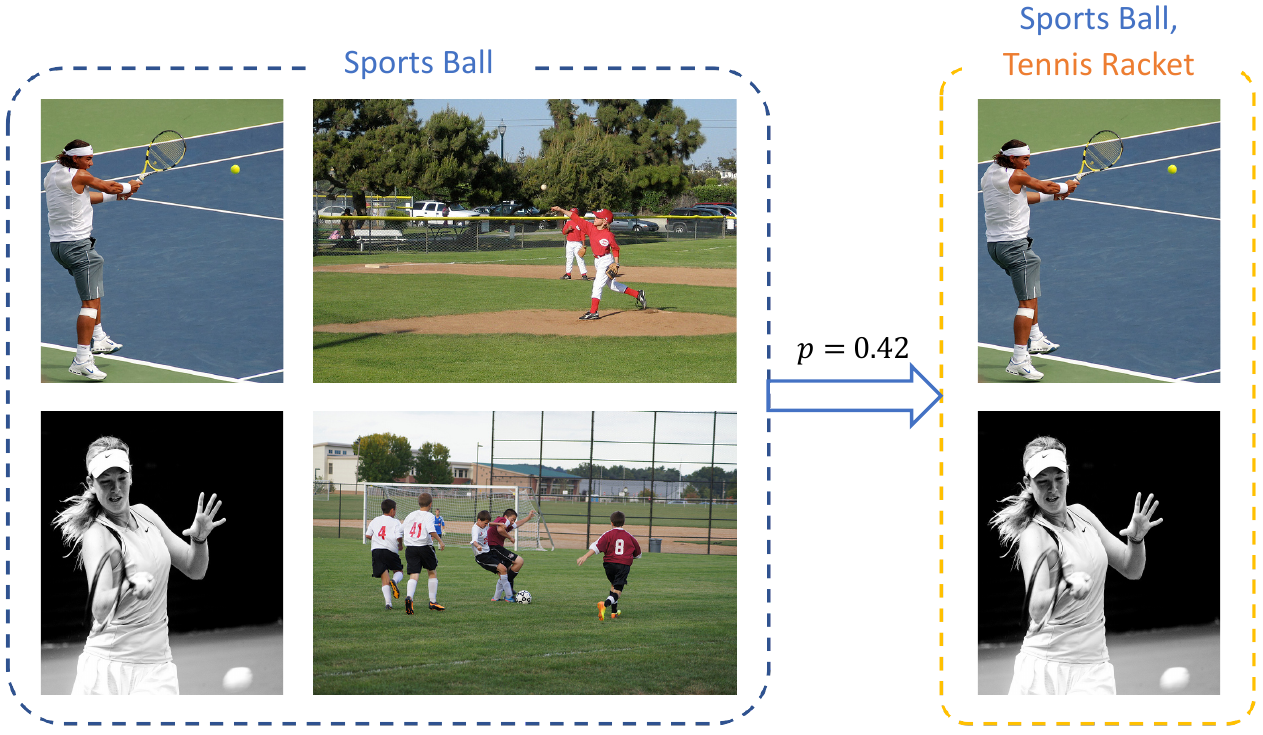} }
	\subfigure[Examples on Charades]  {\includegraphics[width=0.95\columnwidth]{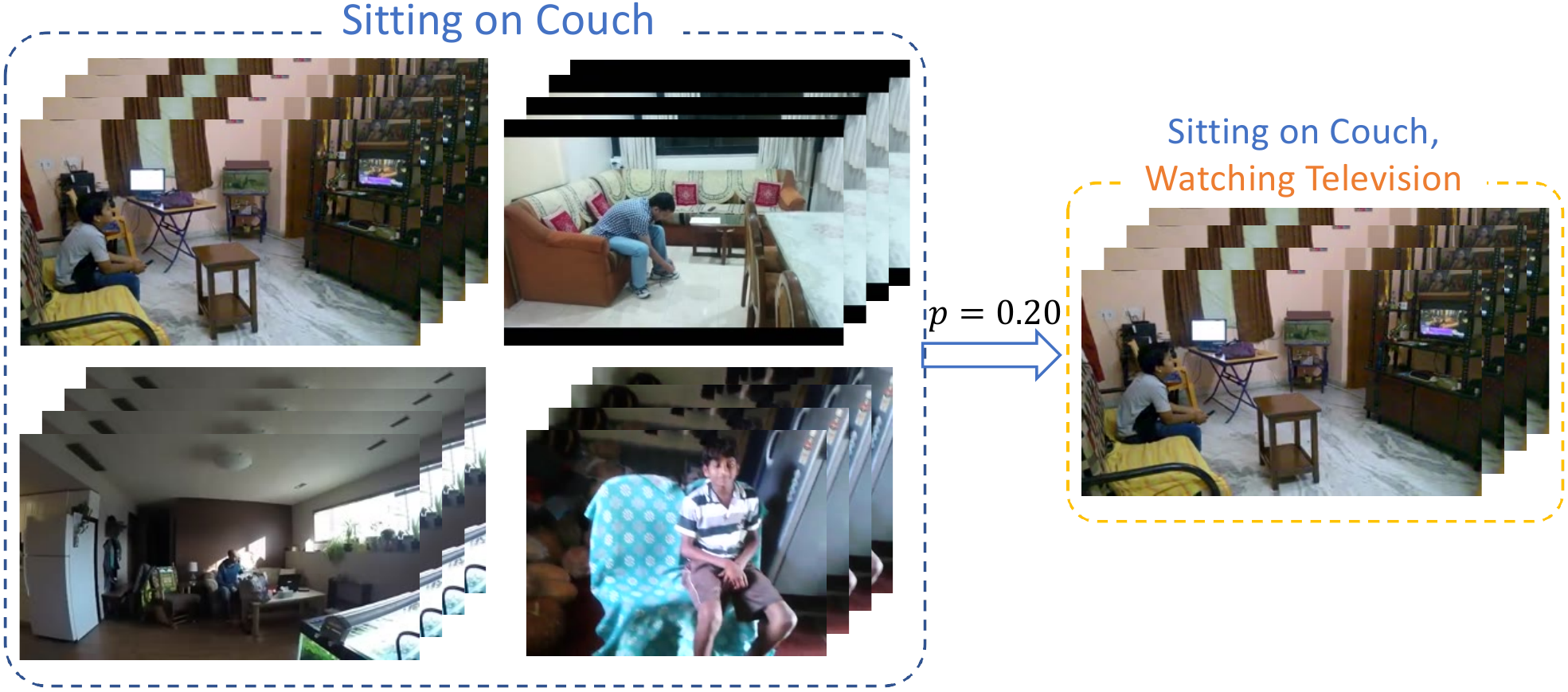} }
	\caption{Examples of label relationship in multi-label datasets. (a) illustrates the co-occurrence of ``Sports Ball'' and ``Tennis Racket'' on the MS-COCO datasets, we can see the frequency that ``Tennis Racket'' co-occurs with ``Sports Ball'' is as high as 0.42. Similarly, (b) showcases an example of ``Sitting on Couch'' and ``Watching Television'' from the Charades dataset.}
	\label{include}
\end{figure}
Multi-label is a natural property of images or videos, it is usually the case that a image or video contains multiple objects or actions. In the computer vision community, multi-label recognition is a fundamental and practical task, and has attracted increasing research efforts. Given the great success of single label image/video classification brought by deep convolutional networks \cite{he2015delving,carreira2017quo,resnet,feichtenhofer2018slowfast,wu2019long}, multi-label recognition can achieve pretty performance by naively treating each label as an independent individual and applying multiple binary classification to predict whether a label presents or not. However, we argue that the following two aspects should be taken into consideration for such a task.

First of all, labels co-occur in images or videos with priors. As illustrated in Figure~\ref{include}, with great chance, ``Sports Ball'' comes together with ``Tennis Racket'' and a man ``Sitting on Couch'' is ``Watching Television'' simultaneously. Then, a question is naturally raised, how to model the relation among labels to leverage such priors for better performance? Secondly, given input $X$, the common practice for predicting its labels can be formulated as a two-stage mapping $y=F_1\circ F_0(X)$, where $F_0:X\mapsto f$ denotes the CNN feature extraction process and $F_1: f\mapsto y$ is the mapping from feature space to label space. Labels are only \emph{explicitly} involved in the last stage as supervision in the training phase. Therefore, the further question is, for a specific multi-label classification task, whether and how the mutual-related label space can explicitly help the feature learning process $F_0$?

To take into account the label correlations, some approaches have been proposed. For example, probabilistic graph model was used in \cite{li2016conditional,li2014multi} and RNN was used in \cite{wang2016cnn} to capture dependencies among labels. However, probabilistic graph models may suffer from scalability issues given their computational cost. RNN model relies on predefined or learned label sequential order and fails to well capture the global dependencies. Recently, graph convolutional network \cite{kipf2016semi}, \emph{aka} GCN, has witnessed prevailing success in modeling relationship among vertices of a graph. Such a tool was leveraged to model the relation of the label system for multi-label recognition in \cite{chen2019multi}. Meanwhile, the label graph was built simply by utilizing the frequency of label co-occurrence. Another direction is to implicitly model label correlations via local image regions attention, as was done in \cite{wang2017multi,zhu2017learning}. In addition, all the aforementioned solutions follow the conventional practice of two-stage mapping and the whole structure of label system is ignored in learning the feature space.
	
In this paper, we attempt to find possible answers for the two questions. We propose a label graph superimposed deep convolution network called \emph{KSSNet} for this task. The superimposing means the following two folds in our framework: (1) to model the priors of co-occurrence of labels following the GCN paradigm, instead of using statistics of label co-occurrence alone to build the relation graph of the label system, we propose to superimpose knowledge based graph into statistics based graph for constructing the final one. (2) In order to learn better feature representations for a specific multi-label recognition task anchored on its label structures, we design a novel superimposed CNN and GCN network to extract label structure aware descriptors. Specifically, we first construct two adjacency matrices $A_S\in R^{N\times N}$ and $A_K\in R^{N\times N}$ to denote correlation graphs of labels, which is constructed by co-occurrence statistics and a knowledge graph named ConceptNet \cite{speer2017conceptnet} respectively. The initial embedding of all nodes (namely, labels) 
is extracted from ConceptNet. The final adjacency matrix is a superimposed version. Then we apply multi-layer graph convolution on the final superimposed graph to model the label correlation. Besides, different from conventional graph augmented CNN solutions which utilize information of label system at the final recognition stage, we add lateral connections between CNN and GCN at shallow, middle and deep layers to inject information of the label system into backbone CNN for the purpose of labels awareness in feature learning. We have carried out extensive experiments on MS-COCO dataset \cite{Lin2014Microsoft} for multi-label image recognition and Charades \cite{Sigurdsson2016Hollywood} for multi-label video classification. Results show that our solution obtains absolute mAP improvement of 6.4\% and 12.0\% in MS-COCO and Charades with very limited computation cost overhead, when compared to its plain CNN counterpart. Our model achieves new state-of-the-art and outperforms current state-of-the-art solution by 1.3\% and 2.4\% in mAP on MS-COCO and Charades, respectively. 


\section{Related Work}
State-of-the-art image or video classification frameworks \cite{resnet,carreira2017quo,feichtenhofer2018slowfast,stnet,wu2019long} can be directly applied for multi-label classification by replacing the cross-entropy loss with multi-binary classification loss. The straightforward extension leaves label correlation unexplored thus degrading the recognition performance. We propose our solution to alleviate this problem and it is closely related with the following jobs.

Many existing works on multi-label classification proposed to capture label relationship for performance improvement.
The co-occurrence of labels can be well formulated by probabilistic graph models, in the literature, there have many methods based on such mathematical theory to model the labels \cite{li2016conditional,li2014multi}. To tackle the problem of computation cost burden of probabilistic graph models, the neural network based solution is becoming prevalence recently.
In \cite{wang2016cnn}, recurrent network was used to encode labels into embedding vectors for label correlation modeling purpose. Context gating strategy was utilized in \cite{lin2018nextvlad} to integrate the post processing of label re-ranking into the whole network architecture. There are also works done by leveraging the attention mechanism in order for modeling label relationship. In \cite{wang2017multi} and \cite{zhu2017learning}, either image region-level spatial attention map or attentive semantic-level label correlation modeling was used to boost the final recognition performance. \cite{wang2019baseline} proposed to improve the performance by model ensemble.

\begin{figure*}[!t]
	\centering
	\includegraphics[width=0.9\textwidth]{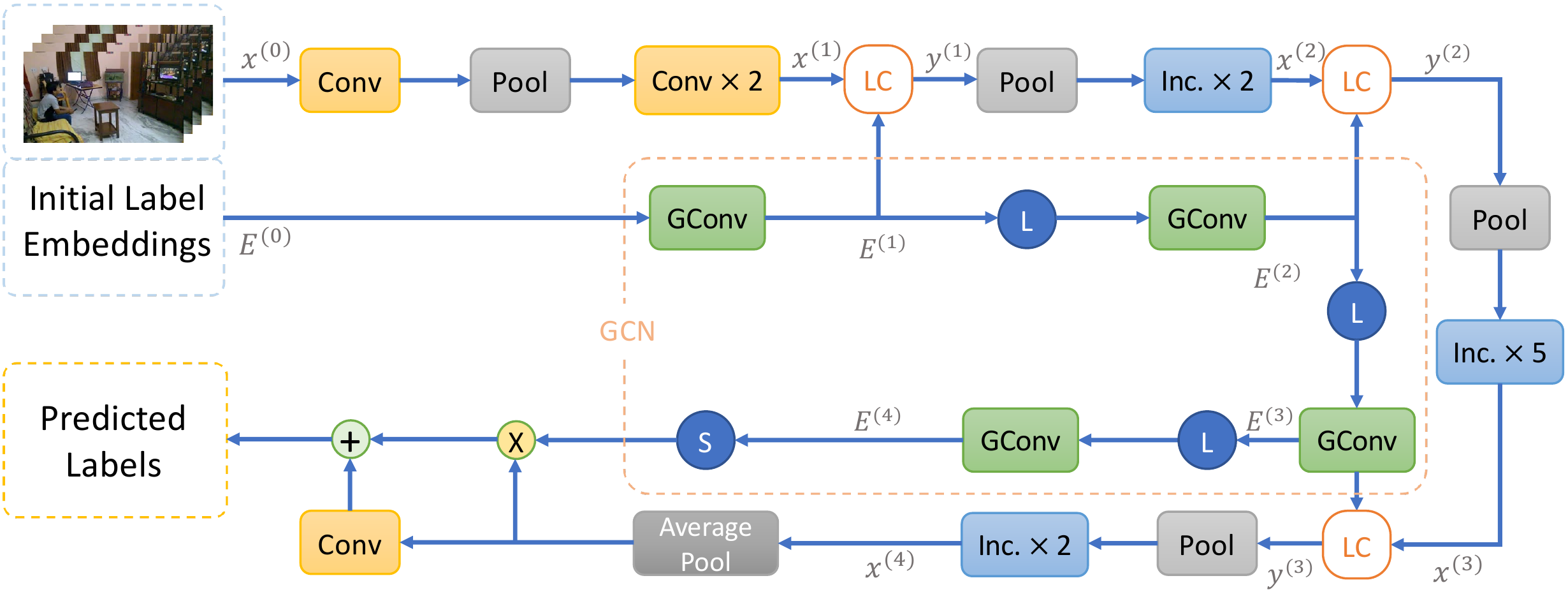}
	\caption{The overview of KSSNet with backbone of Inception-I3D. ``LC'' is our proposed lateral connection, `S' and `L' denote Sigmoid and LeakyReLU operations, respectively. ``Inc.'' is the Inception block in I3D \cite{carreira2017quo}. KSSNet takes videos and initial label embeddings as input, and outputs the predicted labels of these videos. ``GConv'' is the abbreviation of ``Graph Convolution''.}
	\label{sketch}
  \end{figure*}
Graph has been proved to be more effective for label structure modeling. Tree-structure label graph built with maximum spanning tree algorithm in \cite{li2014multi} and knowledge graph for describing label dependency in \cite{lee2018multi} are two typical label graph solutions. Recently, GCN was introduced in \cite{kipf2016semi} and it has been successfully utilized for non-grid structured data modeling. Researchers have leveraged GCN for many computer vision tasks and great performance was achieved. For instance, it was leveraged in \cite{yan2018spatial,gao2018generalized} to model the relationship of skeletons of humans bodies for human action recognition and knowledge-aware GCN was applied for zero-shot video classification in \cite{gao2019know}. Our work mostly relates to the one proposed in \cite{chen2019multi}, which used GCN to propagate information among labels and merges label information with CNN features at the final classification stage. Differently, our work builds GCN by superimposing the graph built from statistical co-occurrence information into the graph built with knowledge priors. The label information is absorbed into the backbone network for better feature learning. 

\section{Approach}
In this paper, We propose a knowledge and label graph superimposing framework for multi-label classification. We provide a new label correlation modeling method of superimposing statistical label graph and knowledge prior oriented label graph. Better feature learning network architecture by absorbing label structure information generated by GCN at shallow, middle and deep layers of backbone CNN is designed. We call our model as \emph{KSSNet}~(Knowledge and Statistics Superimposing Network). Taking the KSSNet with backbone of Inception-I3D \cite{carreira2017quo} designed for multi-label video classification as example, we show its block-diagram in Figure \ref{sketch}. When it comes to multi-label image classification, the framework can be easily constructed by superimposing GCN with state-of-the-art 2D CNN such as ResNet \cite{resnet}. 
In the following subsections, we firstly introduce in detail how label graph are constructed and superimposed, and then we show what is our proposed GCN and CNN superimposing.

\subsection{Graph Construction}
Our final graph is constructed by superimposing statistical label graph into knowledge prior oriented graph.
Graph constructed with such statistical information as label co-occurrence frequencies and conditional probabilities of different labels is termed as \emph{statistical graph} in our paper. Statistical information is determined by the distribution of samples in training set.  
The statistical graph can be influenced significantly by noise and disturbance. Meanwhile, knowledge graph, such as ConceptNet \cite{speer2017conceptnet}, is built with human knowledge by several methods, such as expert-created resources and games with a purpose. It is more authentic for representing the relationship of labels, especially for small scale datasets. However it has three drawbacks: Firstly, the graph is so dense that it represents too much trivial relationship of nodes. When used into deeper GCNs, it will result in more heavy negative effect of over-smoothed label embeddings, compared with sparse graphs. Secondly, it is datasets independent and neglects the characteristics of specific tasks. Thirdly, as knowledge graph can hardly contain all labels in a dataset, the edges of these labels are lost. Our proposed method combines statistical information and human knowledge, which can overcome their drawbacks to some extent. We formally present its details as follows.
\begin{figure}[!tb]
	\centering
	\includegraphics[width=0.8\columnwidth]{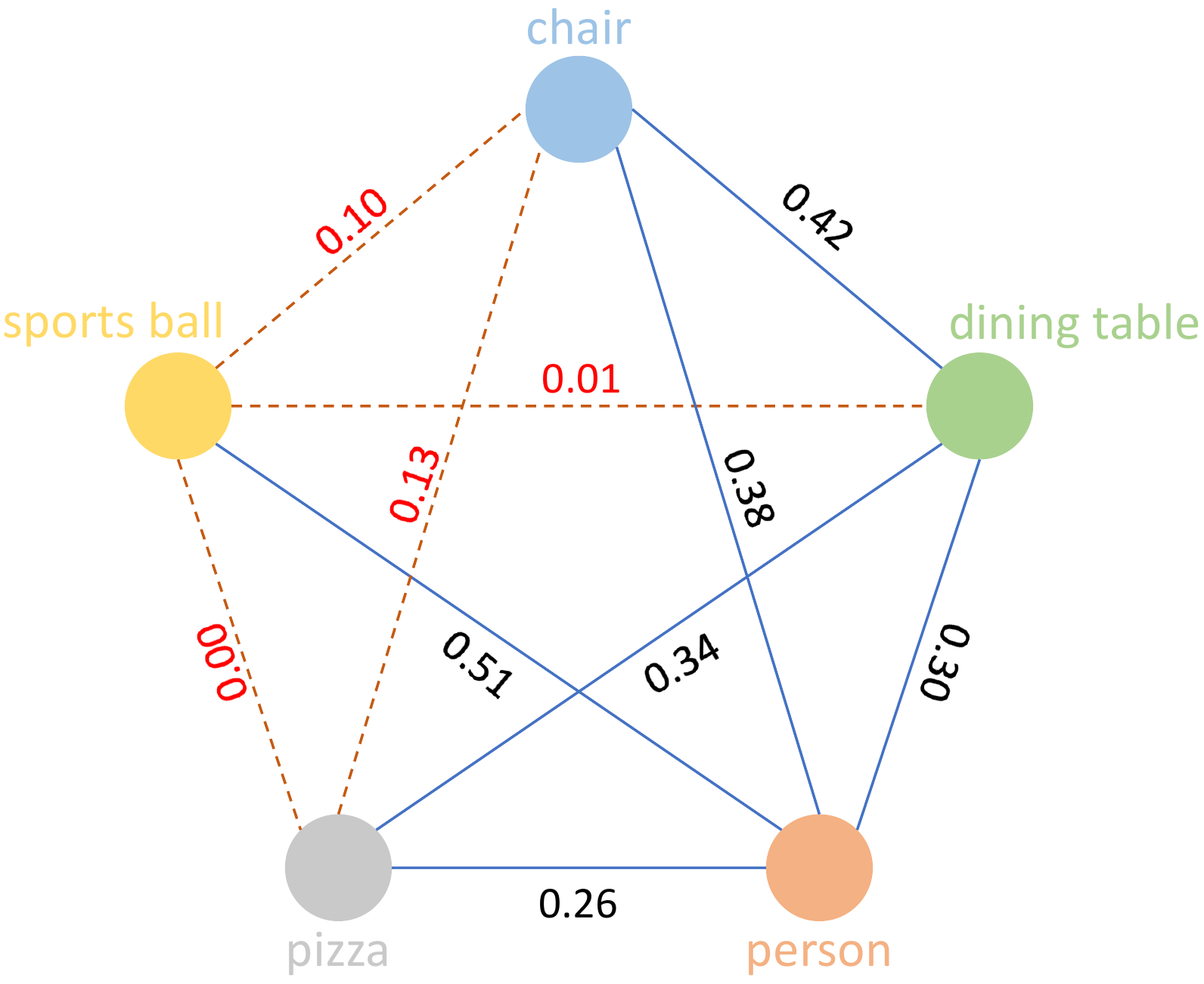}
	\caption{A subgraph with five nodes on MS-COCO.  The number on each edge denotes its weight. Yellow dashed lines with red numbers nearby highlight the redundant edges when taking the threshold of $0.2$.}
	\label{redundant}
  \end{figure}

A graph is usually denoted as $G = (\mathcal{V}, \mathcal{E}, A)$, where $\mathcal{V}$, $\mathcal{E}$, $A$ are the set of nodes, set of edges and adjacency matrix. $A$ is an $N \times N$ matrix with $(i,j)$ entry equaling to the weight of edges between nodes $V_i$ and $V_j$. $N=|\mathcal{V}|$ is the number of vertices. $E\in R^{N \times F}$ denotes the feature  (label embeddings in our case) matrix for all $N$ nodes.


We denote the statistical graph as $G_S = (\mathcal{V}, \mathcal{E}_S, A_S)$, knowledge graph as $G_K = (\mathcal{V}, \mathcal{E}_K, A_K)$, where $A_S$ and $A_K$ are adjacency matrices obtained with statistical information and knowledge priors respectively. $A_S$ is constructed by following \cite{chen2019multi}. $A_K$ is obtained according to the human created knowledge graph ConceptNet \cite{speer2017conceptnet}. Specifically,
\begin{equation}
[A_K]_{ij} = \begin{cases}
\max \{w_r| r \in S_{ij}\}, &  if \ |S_{ij}| > 0\\
0, & if\ |S_{ij}| = 0
\end{cases}
\end{equation}
where $S_{ij}$ is a set of relations (such as ``used for'' and ``is a'') between nodes $\mathcal{V}_i$ and $\mathcal{V}_j$ extracted from ConceptNet. $w_r$ is the weight of relation $r$. $|S_{ij}|$ is the number of elements in $S_{ij}$.


    Denoting $A_S'$ and $A_K'$ as the normalized versions of $A_S$ and $A_K$, respectively. The normalized $A_S$ is $A_S' = D_S^{-1/2} A_S D_S^{-1/2}$, where $D_S$ is diagonal and $[D_S]_{ii} = \sum_j [A_S]_{ij}$. $A_S$ is normalized analogously. Weighted average of $A_S'$ and $A_K'$ is used to superimpose the prior knowledge into statistical graph and the resulted new adjacency matrix is normalized. 
\begin{equation} \label{merge}
	A = \lambda A_S' + (1 - \lambda) A_K'
\end{equation}
where $\lambda \in [0, 1]$ is a weight coefficient.

Meanwhile, as the elements of $A_S'$ and $A_K'$ are non-negative, $A$ has more nonzero elements compared with $A_S$ and $A_K$. That is, the graph constructed with $A$ has more redundant edges than $G_S$ or $G_K$, as is illustrated in Figure \ref{redundant}. In order to suppress these edges, we use a threshold $\tau \in R$ to filter the elements of $A$
\begin{equation} \label{tau}
	[A_{\tau}]_{ij} = 
	\begin{cases}
		0, &if \ A_{ij} < \tau \\
		A_{ij}, &if \ A_{ij} \geq \tau
	\end{cases}
\end{equation}

As is known to us, when the number of GCN layers increases, the performance of models drops in some tasks. The reason is possibly the over-smoothing of deeper GCN layers \cite{chen2019multi}. Inspired by such fact, we further adjust the entries in the adjacency matrix of the superimposed graph and obtain the final matrix $A_{KS}$:
\begin{equation} \label{threshold}
	A_{KS} = \eta A_{\tau} + (1 - \eta) \mathit{I}_N
\end{equation}
where $\mathit{I}_N$ is an $N \times N$ identity matrix. $\eta \in R$ is a weight coefficient.
With the adjacency matrix $A_{KS}$, we construct the set of edges as
\begin{equation}
	\mathcal{E}_{KS} = \{(V_i, V_j) \vert [A_{KS}]_{ij} \neq 0, \ and\  0 \leq i, j \leq N\}
\end{equation}
$(V_i, V_j)$ denotes the edge (directed or undirected) of nodes $V_i$ and $V_j$. The graph we proposed is defined as $G_{KS} = (\mathcal{V}, \mathcal{E}_{KS}, A_{KS})$, which is called \emph{KS graph}.








\subsection{Superimposing of GCN and CNN}
Unlike conventional convolutions, GCN is designed for non-Euclidean topological structure. In GCN, the label embeddings of each node is a mixture of the embeddings of its neighbors from the previous layer. We follow a common practice as was done in \cite{kipf2016semi,chen2019multi} to apply graph convolution. Every GCN layer can be formulated as a non-linear function: 
\begin{equation} \label{eq:gcn0}
	E^{(l+1)} = \sigma(A_{KS}'E^{(l)}W^{(l)}),
\end{equation}
where $A_{KS}'$ is the normalized adjacency matrix. $E^{(l)}\in R^{N \times C^{(l)}}$ denotes the label embedding at the $l$-th layer for all $N$ nodes. Note that $E^{(0)}$ is the initial label embeddings and it is extracted from semantic networks like ConceptNet \cite{speer2017conceptnet}. $W^{(l)} \in R^{C^{(l)} \times C^{(l+1)}}$ is a transformation matrix and is learnable in the training phase. $\sigma(\cdot)$ denotes a non-linear activation operation.

\begin{figure}
	\centering
	\includegraphics[width=1.0\columnwidth]{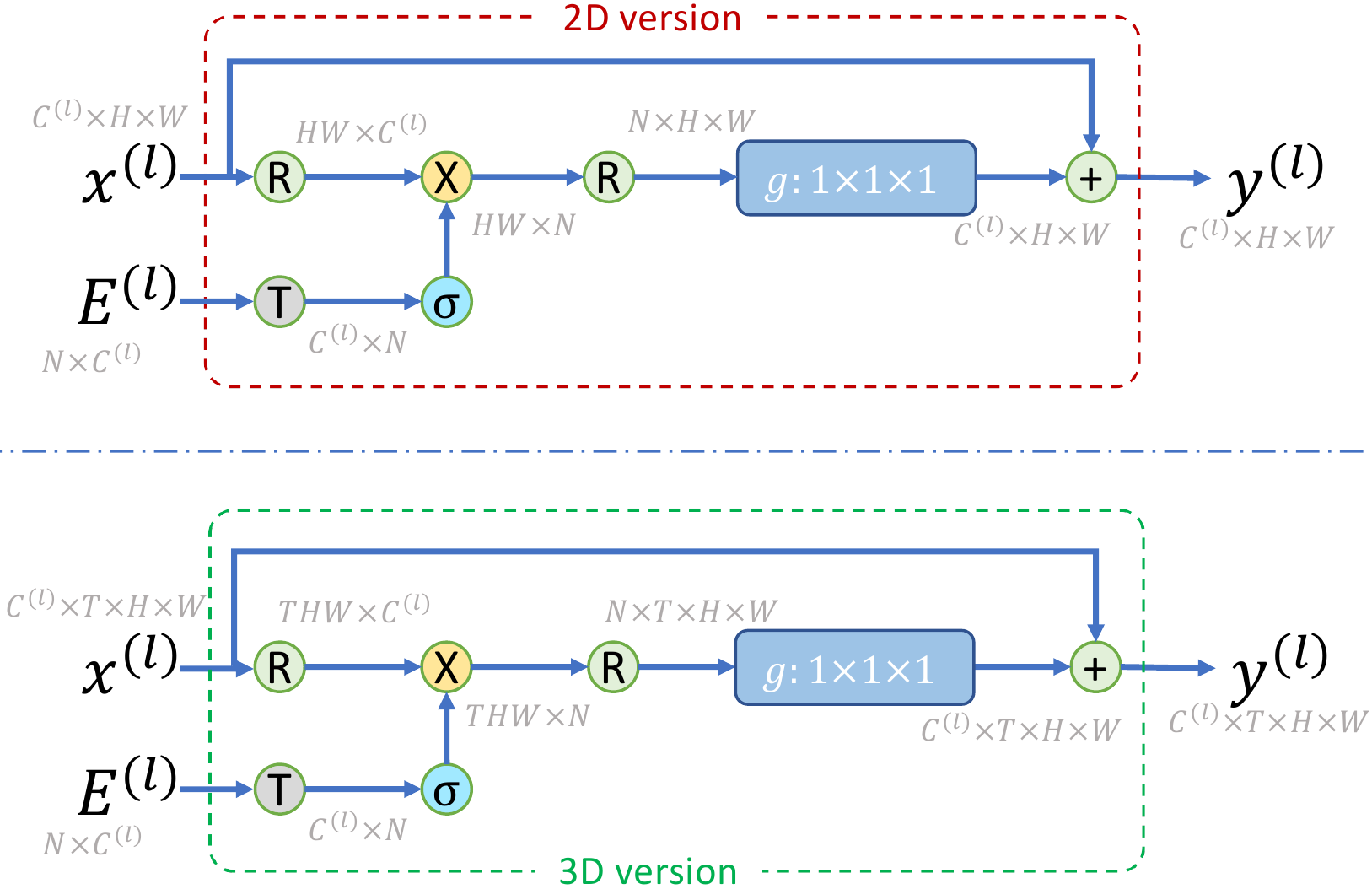}
	\caption{The block diagram of LC operation. `R', ``$(\cdot)^T$'', `$\times$' and `$+$' denote matrix reshape, transpose, multiplication and sum operations respectively. $x^{(l)}$ and $E^{(l)}$ are CNN feature and GCN feature at the $l^{\text{th}}$ GCN layer. The shape of each tensor is marked in gray annotation.}
	\label{connect}
\end{figure}

Instead of only superimposing information of label relationship at the final recognition stage, we propose to inject label information into backbone 2D/3D CNNs at different stages by lateral connection (\emph{LC operation}).
Figure~\ref{connect} shows 2D and 3D versions of our proposed LC operation. Take 3D version for example, we define an LC operation in deep neural networks as:
\begin{equation}\label{eq:lc}
y = g(R_{N\times T\times H \times W}(R_{THW \times C}(x) \otimes \sigma(E^T))) + x
\end{equation}
Here $x \in R^{C\times T\times H \times W}$ is CNN feature, $C$ is the number of channels. $T$, $H$ and $W$ denote the frames, height and width of feature tensor. $N$ is the number of labels. $E \in R^{N \times C}$ indicates the hidden label embeddings of GCN. $g$ is a $1\times 1\times 1$ convolution $g: R^{N \times T\times H\times W} \mapsto R^{C \times T\times H\times W}$, whose parameters are to be learnt for the downstream tasks. `$\otimes$' denotes matrix multiplication and $(\cdot)^T$ is transpose operation. $\sigma(\cdot)$ denotes a non-linear activation operation. Both $R_{N\times T\times H\times W}(\cdot)$ and $R_{THW \times C}(\cdot)$ are defined as reshape operations, which rearrange the input array as the shape noted at their subscripts.

The motivation of LC is to push the CNN network to learn label-system anchored feature representations for better recognition. As stated in (\ref{eq:lc}), it first calculates cross-correlation of CNN features and label embeddings and outputs how each CNN feature point is correlated with a label embedding. Such correlation tensor is then mapped to a hidden space by 1x1x1 convolution to encode the relationship of CNN features and label embeddings. At last, the relationship tensor generated from $1\times1\times1$ convolution are added into the original CNN feature tensor. With the lateral connection, the relationship of label system and CNN feature maps is modeled and the learned CNN feature is kind of label-system anchored.


\begin{table*}[!t]
	\centering
	\footnotesize
	\caption{Performance comparisons between baselines and KSSNet on MS-COCO. KSSNet is based on our proposed KS graph and has four GCN layers.}
	\resizebox{0.85\textwidth}{!}{
	\begin{tabular}{c|c|c|c|c|c|c|c}
	\hline
	{Method} & {mAP} & {CP} & {CR} & {CF1} & {OP} & {OR} & {OF1} \\
	\hline
	CNN-RNN~\cite{wang2016cnn} & 61.2&- &- &- &- &- &- \\
	SRN~\cite{zhu2017learning} & 77.1 & {81.6} & 65.4 & 71.2 & 82.7 & 69.9 & 75.8 \\
	ResNet101~\cite{he2016deep} & 77.3 & 80.2 & 66.7 & 72.8 & 83.9 & 70.8 & 76.8 \\
	Multi-Evidence~\cite{ge2018multi} & -- & 80.4 & 70.2 & 74.9 & 85.2 & 72.5 & 78.4 \\
	ML-GCN~\cite{chen2019multi}  & 82.4 & 84.4 & 71.4 & \textbf{77.4} & 85.8 & 74.5 & 79.8 \\
	\hline
	KSSNet & \textbf{83.7} & \textbf{84.6} & \textbf{73.2} & 77.2 & \textbf{87.8} & \textbf{76.2} & \textbf{81.5} \\
	\hline
	\end{tabular}}
	\label{coco}
	\end{table*}
Our KSSNet superimposes labels embeddings into CNN features not only in the classification layer but also in hidden layers. There are several advantages of this strategy. (1) The hidden embeddings in GCN can help the feature learning process of CNN, making hidden CNN features aware of label relationship. (2) As for the learning process of hidden embeddings, the extra gradients from LC operation can be seen as a special regularization, which forces hidden embeddings more adapt to CNN features. It can overcome the over-smoothing of deeper GCN to some extent.


\section{Experiment}
In this section, we conduct experiments to show that our proposed solution can achieve pretty good performance in both image and video multi-label recognition tasks. Then, we carry out ablation studies to evaluate the effectiveness of the proposed graph construction method in our KSSNet. 

\subsection{Datasets and Evaluation Metrics}

\subsubsection{MS-COCO}
MS-COCO \cite{Lin2014Microsoft} is a static image dataset, which is widely used for many tasks, such as multi-label image recognition, object localization and semantic segmentation. It contains about 82K images for training, 41K for validation and 41K for test. All images are involved with 80 object labels in the multi-label image recognition task. On average, each image has 2.9 labels. 
We evaluate all the methods on validation set, since the ground-truth labels of the test set are not available. 

\subsubsection{Charades}
Charades \cite{Sigurdsson2016Hollywood} is a multi-label video dataset containing around 9.8K videos, among which about 8K for training and 1.8K for validation. The average length of videos in Charades is about 30 seconds. It has 157 action labels and 66.5K annotated activities, about 6.8 labels per video. Each action label is composed of a noun~(object) and a verb~(action). In total, there are 38 different nouns and 33 different verbs. We also evaluate different methos using its validation set.

\subsubsection{Evaluation Metrics}
In order for evaluating our model on MS-COCO comprehensively and for convenience of comparison with other solutions, we report the average per-class precision~(CP), recall~ (CR), F1~(CF1), the average overall precision~(OP), overall recall~(OR), overall F1~(OF1) and mean average precision~(mAP), as is done in \cite{chen2019multi}. 
As for the Charades, we evaluate all models with mAP~\cite{Sigurdsson2016Hollywood} to show their effectiveness. Besides, we also report the value of FLOPs that each model consumes to depict model complexity.

\subsection{Implementation Details}
\subsubsection{Experiment on MS-COCO}
For image recognition, we choose state-of-the-art ResNet101~\cite{he2016deep} as the backbone of our KSSNet, which is pre-trained on ImageNet. The GCN of KSSNet is built from four successive graph convolution layers and the number of channels of their outputs is $256$, $512$, $1024$ and $2048$, respectively. In order to deal with the ``dead ReLU'' problem, we use LeakyReLU as activation operation for graph convolution layers, with negative slope of $0.2$. Three 2D version LC operations between GCN and the backbone ResNet101 are used and the label embeddings of four graph convolution layers are injected to res2, res3, res4 and res5 of ResNet101. The activation function in the LC operation is set to be Tanh.

We adopt $300$-dimentional GloVe text model~\cite{pennington2014glove} to generate the initial label embeddings of labels. As for the labels whose names contain multiple words and have no corresponding keys in GloVe, we obtain the label representation by averaging embeddings of all words. In the process of constructing statistical matrix $G_S$, we use the strategy proposed in \cite{chen2019multi}. We set $\lambda$ in (\ref{merge}) to be 0.4, $\tau$ in (\ref{tau}) to be 0.02 and $\eta$ in (\ref{threshold}) to be 0.4. 
During training, the same data preprocessing procedure as \cite{chen2019multi} is adopted. 
Adam is used as the optimizer with a momentum of 0.9, weight decay of $10^{-4}$ and batch size of 80. The initial learning rate of Adam is 0.01. All models are trained for 100 epochs in total. 

\begin{table*}[t]
\centering
\footnotesize
\caption{Quantitative results of baselines and KSSNet on Charades validation set. The KSSNet bellow has 4 GCN layers and its adjacency matrix is from our proposed KS graph.}
\resizebox{\textwidth}{!}{
\begin{tabular}{c|c|c|c|c|c}
\hline
Method & {Backbone}&{Modality} &{Pretrain} & {mAP}    & {GFLOPs} \\
\hline
Two-stream~\cite{wu2018compressed} &VGG16 &RGB+Flow &  ImageNet,UCF101 &  14.3 & -- \\
CoViAR~\cite{wu2018compressed}  &--&Compressed~\cite{wu2018compressed} &   ILSVRC2012-CLS &  21.9 & -- \\
CoViAR~\cite{wu2018compressed}&-- &Compressed+Flow &   ILSVRC2012-CLS &  24.1 & -- \\
Asyn-TF~\cite{Sigurdsson2017Asynchronous} &VGG16 &RGB+Flow &  ImageNet  &  22.4 & --  \\
MultiScale~(TR)~\cite{zhou2018temporal} &Inception-I3D&RGB &  ImageNet  &  25.2 & -- \\
I3D~\cite{carreira2017quo} & Inception &RGB & Kinetics-400 & 32.9 &  \textbf{108}\\
ResNet-101(NL)~\cite{wang2018non} &ResNet101-I3D &RGB &  Kinetics-400  &  37.5 & 544  \\
STRG~(NL)~\cite{wang2018videos} &ResNet101-I3D&RGB &  Kinetics-400 &   39.7 & 630  \\
SlowFast~\cite{feichtenhofer2018slowfast} &ResNet101&RGB & Kinetics-400 & 42.1 & 213 \\
SlowFast(NL)~\cite{feichtenhofer2018slowfast} &ResNet101&RGB & Kinetics-400 & 42.5  &  234  \\
LFB(NL)~\cite{wu2019long} &ResNet101-I3D &RGB &  Kinetics-400 & 42.5 & -- \\
\hline
KSSNet&  Inception-I3D &RGB & ImageNet & \textbf{44.9} & 127 \\
\hline
\end{tabular}}
\label{charades}
\end{table*}

\subsubsection{Experiment on Charades}
Inception-I3D of KSSNet is initialized following the inflating mechanism proposed in I3D \cite{carreira2017quo} with BN-Inception pretrained on ImageNet. We fine-tune our models using 64-frame input clips. These clips are sampled following the strategy of \cite{wang2016temporal}, where each clip consists of 64 snippets and each snippet contain only one frame. The spatial size is $224\times 224$, randomly cropped from a scaled video whose spatial size is $256\times 256$. $\lambda$, $\eta$ and $\tau$ are set to 0.6, 0.4 and 0.03, respectively. We train all models with mini-batch size of 16 clips. Adam is used as the optimizer, starting with a momentum of 0.9 and weight decay of $10^{-4}$. The weight decays of all bias are set to zero. Dropout~\cite{hinton2012improving} with a ratio of 0.5 is added after the average pooled CNN features. The initial learning rate of GCN parameters is set to be 0.001, while others are set to be $10^{-4}$. We use the strategy proposed in \cite{he2015delving} to initialize the GCN and initial label embeddings are extracted with ConceptNet \cite{speer2017conceptnet}.
During inference, we evenly extract 64 frames from the original full-length video.

\subsection{Comparison with Baselines}
In this part, we present comparisons with several state-of-the-arts on MS-COCO and Charades, respectively to show the effectiveness of our proposed solution.

\subsubsection{Results on MS-COCO}  
We compare our KSSNet with the state-of-the-art methods, including CNN-RNN~\cite{wang2016cnn}, SRN~\cite{Zhu_2017_CVPR}, ResNet101~\cite{he2016deep}, Multi-Evidence~\cite{ge2018multi} and ML-GCN~\cite{chen2019multi}. Table~\ref{coco} records the quantitative results of all models on MS-COCO validation set. ML-GCN is a GCN+CNN framework based on statistical label graph and it is the current state-of-the-art. It can be observed that our KSSNet obtains the best performance at almost all evaluation matrices. Specially, compared with ML-GCN, its mAP is 1.3\% higher, the improvement of overall precision is improved from 85.8\% to 87.8\%, the gain of overall recall is 1.7\% and new state-of-the-art overall F1 score of 81.5\% is achieved. The result demonstrates the effectiveness of our KSSNet framework. The comparison of KSSNet and its backbone ResNet101 shows that the absolute improvement in mAP is up to 6.4\% and evidences that the label embeddings of GCN can explicitly take advantage of the label relationship, which is hard to be learned by plain CNN or even ignored by many frameworks.

\subsubsection{Results on Charades}

Table~\ref{charades} shows the comparison with state-of-the-art models for our proposed KSSNet on Charades. Compared with backbone I3D model, KSSNet provides 12.0\% higher mAP at the cost of very little computation overhead (from 108 GFLOPs to 127 GFLOPs). We can see that the gain is even larger than what achieved on MS-COCO (12.0\% v.s. 6.4\%). The origin beneath is possibly the characteristics of Charades dataset. On the one hand, each video has 6.8 labels on average, which is even more than MS-COCO. The correlation among different labels has significant influence on multi-label video recognition task. On the other hand, the dataset is not sufficiently large, so the impact of extra label correlation introduced by GCN is more obvious. It can be concluded from such observation that our proposed GCN and CNN superimposing framework can significantly improve baseline result, especially when the training data is not so sufficient. 
We can also see that although no pretraining on extra large scale video dataset, KSSNet~(KS graph) achieves the best performance, which is 2.4\% higher than the current state-of-the-art method LFB and SlowFast(NL) which are pretrained on Kinetics-400. It should be noted that the GFLOPs of our KSSNet is much smaller than SlowFast(NL), which means KSSNet has remarkable potential in fast multi-label video classification.

\begin{table*}[t]
	\centering
	\footnotesize
	\caption{Performance comparisons of different label graphs on MS-COCO and Charades. ``KSSNet~(statistical graph)'', ``KSSNet~(knowledge graph)'' and ``KSSNet~(KS graph)'' are three versions of our proposed KSSNet which have the same framework and different graphs on each dataset. All our variants have four GCN layers.}
	\resizebox{\textwidth}{!}{
	\begin{tabular}{c|c|c|c|c|c|c|c|c|c|c}
	\hline
	\multirow{2}{*}{{Methods}} & \multicolumn{8}{c|}{{MS-COCO}} & \multicolumn{2}{c}{{Charades}} \\
	\cline{2-11}
	&{Backbone}& {mAP} & {CP} & {CR} & {CF1} & {OP} & {OR} & {OF1} &{Backbone}& {mAP} \\
	\hline
	KSSNet~(statistical graph)&ResNet101 & 83.1 & 84.2 & 73.1 & \textbf{77.6} & 87.2 & 76.4 & 81.2 &Inception-I3D& 40.7 \\
	KSSNet~(knowledge graph)&ResNet101 & 81.0 & 82.8 & 69.5 & 75.6 & 84.5 & 73.4 & 78.6 &Inception-I3D& 41.1 \\
	KSSNet~(KS graph)&ResNet101 & \textbf{83.7} & \textbf{84.6} & \textbf{73.2} & 77.2 & \textbf{87.8} & \textbf{76.2} & \textbf{81.5}&Inception-I3D & \textbf{44.9}  \\
	\hline
	\end{tabular}}
	\label{graph-cococharades}
	\end{table*}

\begin{table*}[t]
\centering
\caption{Performance of different GCN depths of KSSNet. On each experiment, all versions of KSSNet use KSS graph as adjacency matrix.}
\resizebox{0.9\textwidth}{!}{
\begin{tabular}{c|c|c|c|c|c|c|c}
	\hline
	\cline{2-8}
	\multirow{2}{*}{{Methods}} & \multicolumn{3}{c|}{{MS-COCO}} & \multicolumn{4}{c}{{Charades}} \\
	\cline{2-8}
	 &{Backbone} & {mAP}  & {\#Params} &{Backbone} & {mAP} & {GFLOPs} & {\#Params}\\
	\hline
	KSSNet~(2 layers)&ResNet101 & 82.9 & \textbf{172.1MB} &Inception-I3D& 41.9 & \textbf{110} & \textbf{47.6MB}\\
	KSSNet~(3 layers) &ResNet101& 83.5  & 173.2MB  &Inception-I3D& 43.8 & 115 & 49.1MB\\
	KSSNet~(4 layers) &ResNet101& \textbf{83.7}  & 173.8MB&Inception-I3D& \textbf{44.9} & 127 & 49.8MB \\
	\hline
\end{tabular}}
\label{layers}
\end{table*}
\subsection{Ablation Studies}
In this section, we perform ablation studies to evaluate the effectiveness of our KS graph and to analyze the influence of GCN depth in KSSNet framework. 

\subsubsection{Label graphs of KSSNet}
In order to evaluate the influence of different graph, we implement three versions of KSSNet with statistical graph, knowledge graph and our proposed KS graph. All of them have the same framework with four GCN layers. 
Table~\ref{graph-cococharades} summarizes the results of KSSNet~(statistical graph), KSSNet~(knowledge graph) and KSSNet~(KS graph). The experiment on MS-COCO shows that knowledge graph performs worse than statistical graph and KS graph, which is caused by the relationship missing of uncovered labels in knowledge graph and by the over-smoothing impact introduced by the presence of many trivial edges. 
However, the experiment on Charades exhibits a contrary result, KSSNet~(knowledge graph) outperforms KSSNet~(statistical graph) by a mAP of 0.4. The cause can attribute to the characteristics of Charades. In Charades, labels are more complex and training samples are not so sufficient. Graph constructed from co-occurrence information is not so reliable while knowledge priors are always valid, so the contradiction between complex label relationship modeling and the lack of samples in Charades makes knowledge graph more effective than statistical graph. 
Both experiments show that our KS graph performs the best, which validates the effectiveness of superimposing statistical graph and knowledge graph.

\subsubsection{Influence of GCN Depth in KSSNet}
As we know, conventional GCN suffers from over-smoothing. In this part, we conduct multi-label image and video recognition experiments to demonstrate that our KSSNet framework can deal with this problem effectively. 

The backbone of KSSNet is ResNet101 and Inception-I3d for MS-COCO and Charades, respectively. 
In this experiment, we modify the GCN pathway of KSSNet to be with three and two graph convolution layers. The modification can be simply done in two steps: 1) delete the first one or two graph convolution layer(s) and the corresponding LC operation(s) from the GCN pathway; 2) then the first graph convolution layer of the rest ones is adapted to take the initial label embeddings $E^{(0)}$ as input by adjusting its number of input channel $C$ to the channel number of $E^{(0)}$. 

Experimental results are shown in Table~\ref{layers}.  It is obvious, with our KSSNet, more GCN layers lead to better classification results at the cost of small increase of computational cost and model size. KSSNet~(3 layers) achieves better performance than KSSNet~(2 layers) by absolute mAP improvements of 0.6\% and 1.9\% in MS-COCO and Charades. KSSNet~(4 layers) outperforms KSSNet~(3 layers) by 0.2\% and 1.1\% in mAP. As is reported in ML-GCN~\cite{chen2019multi}, when GCN has no less than 2 layers, performance of conventional GCN+CNN solution degrades as long as the number of graph convolution layers gets larger. Our model performs on the contrary. This is because that (1) more GCN layers bring more LC operations which guide CNN to learn better label structure aware features at shallow, middle and higher CNN layers. (2) The extra gradients from LC operation can regularize the learning of label embeddings in GCN. (3) We have proposed such strategies as redundant removal to tackle the over-smoothing issue of GCN. 

\textbf{Impact of Super-Parameters}
This experiment is conducted on Charades. When $\lambda$ varies from 0 to 1 by step of 0.2 and keep other parameters as described above, mAP is 41.05, 41.7, 43.44, 44.93, 44.83 and 41.57. When we fix $\lambda$ as 0.6, $\tau$ varies from 0.01 to 0.04 by step of 0.01, the mAP is 44.6, 44.62, 44.93 and 44.77. 

\section{Conclusion}

Capturing label relationship takes a crucial position on multi-label recognition. In order to better model this information, we propose to construct the KS graph for label correlation modeling by superimposing knowledge graph into statistical graph. Then the LC operation is presented for injecting GCN embeddings into CNN features, resulting in a novel neural network KSSNet. LC operation acts as label-feature correlation modeling and helps the model learn label-anchored feature representations. The KSSNet is proven to be capable of learning better feature representations for a specific multi-label recognition task anchored on its label relationship. Experiments on MS-COCO and Charades have demonstrated the effectiveness of our proposed KS graph and KSSNet for both multi-label image and video recognition tasks.

\paragraph{Acknowledgement} This work was supported by the Joint Laboratory of Intelligent Sports of China Institute of Sport Science (CISS).
	
{	
	\small
	\bibliographystyle{aaai}
	\bibliography{AAAI-WangY.951}
}

\end{document}